\pdfoutput=1

\documentclass[11pt]{article}

\usepackage[]{acl}
\usepackage{times}
\usepackage{latexsym}
\usepackage{amsmath}
\usepackage{amsfonts}
\usepackage[T1]{fontenc}
\usepackage{graphicx}
\usepackage[utf8]{inputenc}
\usepackage{hyperref}
\usepackage{epstopdf} 

\usepackage{multirow}
\usepackage{microtype}
\usepackage[lined,boxed,commentsnumbered]{algorithm2e}
\title{Semantic Structure based Query Graph Prediction for Question Answering over Knowledge Graph}

\author{Mingchen Li \\
  Department of Computer Science\\ 
 Georgia State University, Atlanta, USA\\

  \texttt{mli33@student.gsu.edu} \\
  \And
     	Shihao Ji\\
   Department of Computer Science\\ 
  Georgia State University, Atlanta, USA\\
 \texttt{sji@gsu.edu}

}

\begin{document}
\maketitle

\begin{abstract}
Building query graphs from natural language questions is an important step in complex question answering over knowledge graph (Complex KGQA). In general, a question can be correctly answered if its query graph is built correctly and the right answer is then retrieved by issuing the query graph against the KG. Therefore, this paper focuses on query graph generation from natural language questions. Existing approaches for query graph generation ignore the semantic structure of a question, resulting in a large number of noisy query graph candidates that undermine prediction accuracies. In this paper, we define six semantic structures from common questions in KGQA and develop a novel Structure-BERT to predict the semantic structure of a question. By doing so, we can first filter out noisy candidate query graphs, and then rank the remaining candidates with a BERT-based ranking model. Extensive experiments on two popular benchmarks MetaQA and WebQuestionsSP (WSP) demonstrate the effectiveness of our method as compared to state-of-the-arts. The source code can be found at \url{https://github.com/ToneLi/SSKGQA}.

\end{abstract}


\section{Introduction}
Knowledge graph (KG) is a graph structured database~\cite{miller1995wordnet}, in which nodes represent entities (e.g., \emph{Hedgehog in the Fog}, \emph{Sergei Kozlov}), and edges reflect the relations between entities (e.g., \emph{Hedgehog in the Fog} - \emph{written\_by} - \emph{Sergei Kozlov}). Users can get crisp answers by querying KGs with natural language questions, and this process is called Question Answering over Knowledge Graph (KGQA). Recently, consumer market witnesses a widespread application of this technique in a variety of virtual assistants, such as Apple Siri, Google Home, Amazon Alexa, and Microsoft Cortana, etc.

\begin{figure}[htbp]
	\centering
	\includegraphics[width=0.30\textheight]{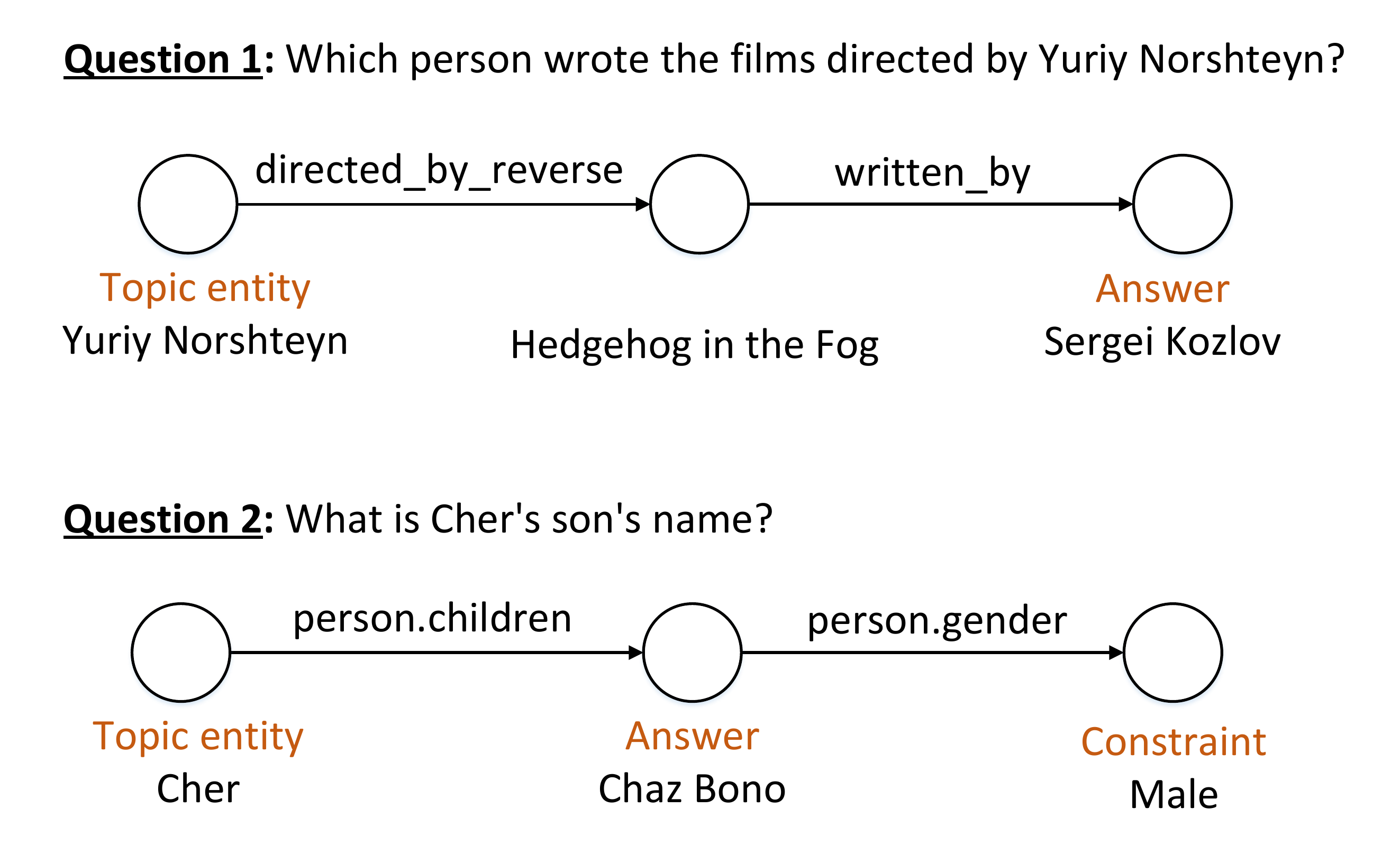}\vspace{-5pt}
	\caption{(Q1) Example question involving multi-hop reasoning, and (Q2) Example question with constraints}\label{fig:complex_questions} 
\end{figure}

Early works~\cite{bordes2015large, golub2016character} on KGQA mainly focus on simple questions, such as \textit{where is the hometown of Obama?} This question only involves in one relation path (e.g.,~\textit{hometown} or~\textit{birth-place}) in KG, and is relatively easy to solve. However, many questions in daily QA sessions are often more complex, manifested by multi-hop reasoning or questions with multiple constraints. Therefore, recently there is a flurry of interests on complex KGQA~\cite{shi2021transfernet,yadati2021knowledge}.

\begin{figure}[tbp]  
	\centering
	\includegraphics[width=0.30\textheight]{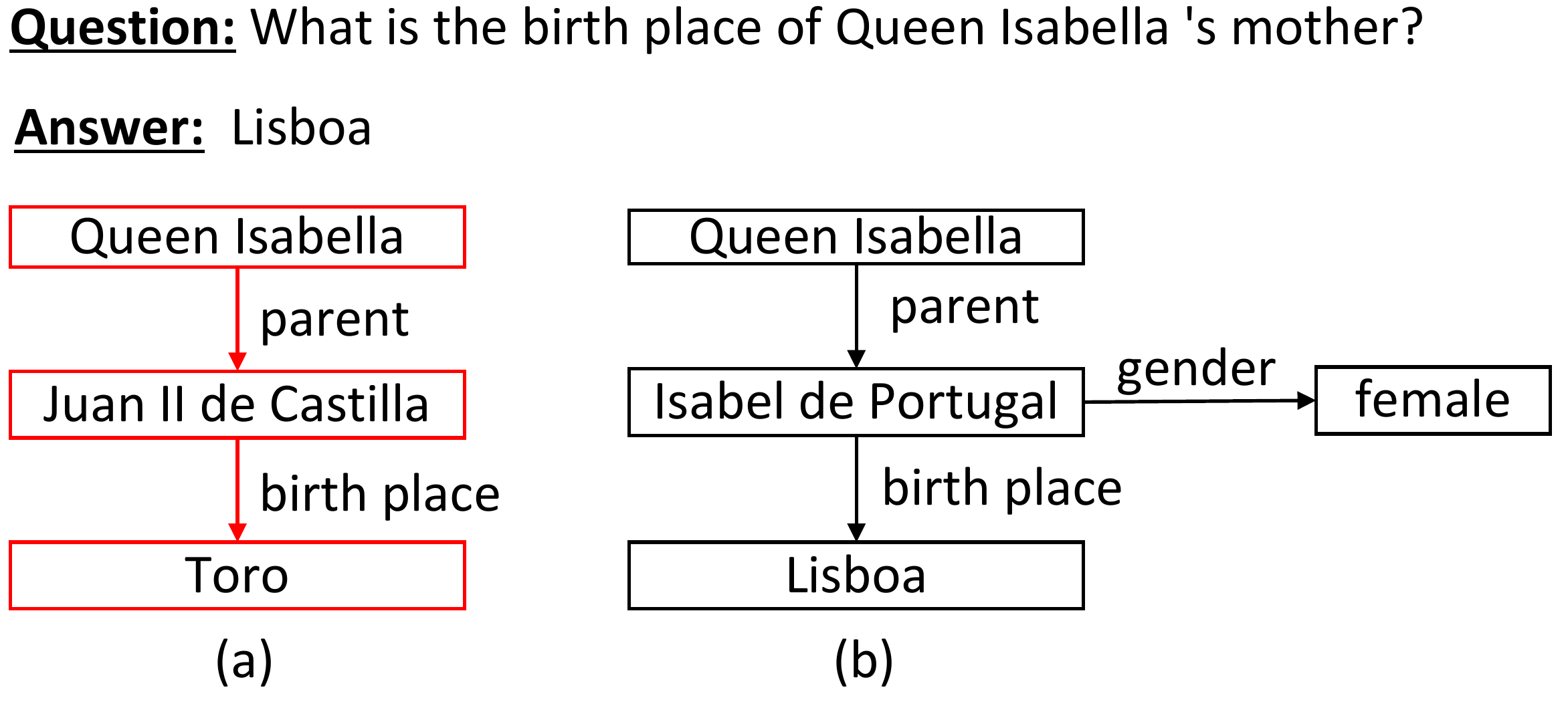}\vspace{-10pt}
	\caption{Example two-hop candidate query graphs for a question with constraints. (a) A candidate query graph with an incorrect semantic structure, (b) A candidate query graph with a correct semantic structure.}\label{fig:query_graph}  \vspace{-10pt}
\end{figure}

There are two types of complexity when dealing with complex KGQA, i.e., multi-hop questions and questions with multiple constraints (See Figure~\ref{fig:complex_questions} for example). Question 1 in Figure~\ref{fig:complex_questions} is a typical multi-hop question, to which the answer is related to \texttt{Yuriy Norshteyn} with two-hop relations: \texttt{directed.by.reverse} and \texttt{written.by}. In response to this challenge, \citet{Kun2019Enhancing} enhances the traditional Key-Value Memory Neural Networks (KV-MemNNs)~\cite{Miller2016Key} for multi-hop question answering. They design a query updating strategy to decompose the question and predict the relevant relation paths at each hop. TransferNet~\cite{shi2021transfernet} is an effective and transparent model for multi-hop questions; it attends to different parts of the question at each hop and computes activated scores for relation path prediction. Despite the promising results, it's still challenging for these models to predict relation paths accurately at each hop, and thus suffer from error propagation over multi-hop reasoning. Similarly, Question 2 in Figure~\ref{fig:complex_questions} is an example of question with constraints. Apparently, there is a single relation path between the topic entity \textit{Cher} and the answer \textit{Chaz Bono}, but the constraint of~\texttt{person.gender=Male} must be satisfied. To handle this type of complex questions, many works built on the idea of query ranking are proposed~\cite{yih2015semantic,lan2020query,chen2020formal}, which rank candidate queries by the similarity scores between question and candidate queries. Specifically, these ranking methods use query graphs to represent queries, and explore various strategies to generate candidate query graphs for ranking. Typical strategies assume the answers are within $n$ hops of topic entity, and enumerate all the relation paths within $n$ hops to generate candidate query graphs. Although this candidate generation strategy can yield all valid query graphs from a topic entity, they have two main limitations: (1) The generated candidate query graphs are very noisy. As shown in Figure~\ref{fig:query_graph}(a), a candidate query graph with an incorrect structure is presented; this candidate query graph is generated by the traditional enumeration strategy but lacks of the constraint on \texttt{person.gender}, which can incur considerable error in query graph ranking (See Table~\ref{ref:with ss}). For the example in Figure~\ref{fig:query_graph}(a), both \texttt{parent} and \texttt{birth.place} are relevant to the question; even though this candidate query graph has an incorrect semantic structure (to be defined in Sec.~\ref{section:SS}), it is still challenging for ranking models to demote it below the correct query graph -- the one in Figure~\ref{fig:query_graph}(b). (2) When building a ranking model to rank query graphs, recent works~\cite{lan2020query} treat the candidate query graph and question as a sequence of words and leverage BERT~\cite{devlin2018bert} to extract feature representation from its pooled output. However, this pooled output is usually not a good representation of the semantics of the input sequence~\cite{khodeir2021bi}. Therefore, improved ranking models are to be developed.


To mitigate the aforementioned issues, this paper proposes \emph{SSKGQA}, a Semantic Structure based framework for complex KGQA. We represent both the multi-hop questions and questions with constraints in a way similar to query graphs\footnote{For example, one candidate query graph for Question~1 in Figure~\ref{fig:complex_questions} can be written as
$Yuriy. Norshteyn \to directed.by.reverse\to y \to written.by \to x
$.} 
and rank the generated candidate query graphs by the similarity scores between question and candidate query graphs. Inspired by~\citet{chen2020formal}, if the structure of a question is known in advance, the noise in candidate query graphs can be reduced significantly by filtering. Thus, SSKGQA first predicts the semantic structure of a natural language question, which is then used to filter out noisy candidate query graphs (which have incorrect structures).

Specifically, we define six semantic structures based on the question topology that is introduced by~\citet{srivastava-etal-2021-complex}. With the defined semantic structures, our SSKGQA processes a natural language question in two stages. In the first stage, we develop a novel Structure-BERT to predict the semantic structure of a natural language question, which is then used to filter out noisy candidate query graphs and produce a set of query graph candidates that match the predicted structure. In the second stage, we rank the remaining candidate query graphs of a question by a BERT-based ranking model and identify the top-1 candidate query graph, which is then issued to retrieve the final answer from a KG. Our experiments demonstrate that this semantic structure based query graph prediction strategy is very effective and enables SSKGQA to outperform state-of-the-art methods.

Our main contributions are summarized as follows. (1) We propose SSKGQA, a semantic structure based method to predict query graphs from natural language questions. SSKGQA can handle both multi-hop questions and questions with constraints and is a unified framework for complex KGQA. (2) We develop a novel Structure-BERT to predict the semantic structure of each question, and a BERT-based ranking model with a triplet loss to identify the top-1 query graph candidate. (3) Compared to state-of-the-arts methods, our SSKGQA demonstrates superior performance on two popular complex KGQA benchmarks.

\section{Related Work}
\subsection{Multi-hop Question Answering}
Current works on multi-hop question answering mainly focus on how to retrieve answers by calculating the relation paths step by step. In general, a right answer can be retrieved if the relation paths are identified correctly at each hop. \citet{Kun2019Enhancing} enhances the traditional Key-Value Memory Neural Networks (KV-MemNNs)~\cite{Miller2016Key} and designs a query updating strategy to decompose the question and predict the relevant relation paths at each hop. TransferNet~\cite{shi2021transfernet} calculates the relation path scores based on an updated question at each hop, but they leverage the attention mechanism to update question representations over multiple hops.
More recently, \citet{cai2021deep} introduces the dual process theory to predict the relation paths at each hop. Although these methods achieve promising results, they suffer from error propagation when predicting the relation paths step by step. To mitigate this issue, SSKGQA identifies the top-1 query graph by directly calculating the similarity scores between question and candidate query graphs (or similarly relation paths).

On the other hand, \citet{Sun2019PullNet,Sun2018Open} incorporate external corpus to enhance the performance of KGQA. They focus on how to get the answers by constructing a subgraph for each question. A challenge of this method is that it is difficult to construct a subgraph around topic entity because we need to identify relevant entities from external corpus, and this process is error-prone. \citet{Apoorv2020Improving} predicts the answers by utilizing the KG embedding model. However, complex questions with long relation paths can reduce the learnability of KG embedding significantly. Our SSKGQA does not need external corpus to improve prediction accuracies and can solve complex multi-hop questions by a semantic structure based ranking.

\subsection{Complex Questions with Constraints}

For questions with constraints, a sequence of works focus on how to reach the answers by generating query graphs. \citet{yih2015semantic} enumerates all possible entities and relation paths that are connected to a topic entity to generate candidate query graphs, and uses a CNN-based ranking model to identify the query graph. Following a similar candidate query graph generation of \cite{yih2015semantic}, \citet{maheshwari2019learning} propose a novel query graph ranking method based on self-attention. \citet{qin2021improving} introduces a query graph generation method by using their proposed relation subgraphs. However, these methods largely ignore the noise when generating the candidate query graphs, which undermines the predictive performance during query graph ranking. To mitigate this issue, SSKGQA first predicts the semantic structure of a question, which is then used to reduce the noise in candidate query graphs.



\subsection{Query Graph Ranking}
Current research on KGQA mainly focuses on how to generate the candidate query graphs, and there are only a few works exploring how to rank the candidate query graphs. \citet{lan2020query} concatenates question and candidate query graph into a single sequence, and leverages BERT \cite{devlin2018bert} to process the whole sequence for ranking. However, \citet{reimers2019sentence} show that this strategy is inefficient as it can incur a massive computation due to the combinatorial nature of concatenation of question and candidate query graphs, leading to duplicated calculation. \citet{chen2020formal} explore GRUs to encode the question and query graph information, and utilize a hinge loss to learn a ranking model. However, GRUs can only learn a limited interaction among words in a sentence, while the global interactions among words has proven to be critical for text representation in various NLP applications~\cite{khan2020mmft}. To solve the aforementioned issues, SSKGQA exploits separated BERT models to process questions and query graphs, respectively, and reuses the extracted features to avoid duplicated calculation and leverages a triplet loss to train a BERT-based ranking model.

\section{Preliminaries}

\subsection{Query Graph}
\label{query graph}
Query graph is a graph representation of a natural language question~\cite{yih2015semantic}. See Figure~\ref{fig:query_graph} for example. A query graph usually contains four types of components: (1) a grounded entity, which is an entity in KG and is often the topic entity of a question, e.g., \texttt{Queen Isabella} in Figure~\ref{fig:query_graph}. (2) an existential variable, which is an ungrounded entity, e.g., \texttt{Isabel de Portugal} in Figure~\ref{fig:query_graph}. (3) a lambda variable, which is the answer to question but usually an ungrounded entity, e.g., \texttt{Lisboa} in Figure~\ref{fig:query_graph}. (4) some constraints on a set of entities, e.g., \texttt{gender} in Figure~\ref{fig:query_graph}. A question can be correctly answered if its query graph is built correctly and the right answer can be retrieved by issuing the query graph (represented by a SPARQL~\cite{perez2009semantics} command) to a KG.

\begin{figure}[t]  
	\centering
	\includegraphics[width=0.3\textheight]{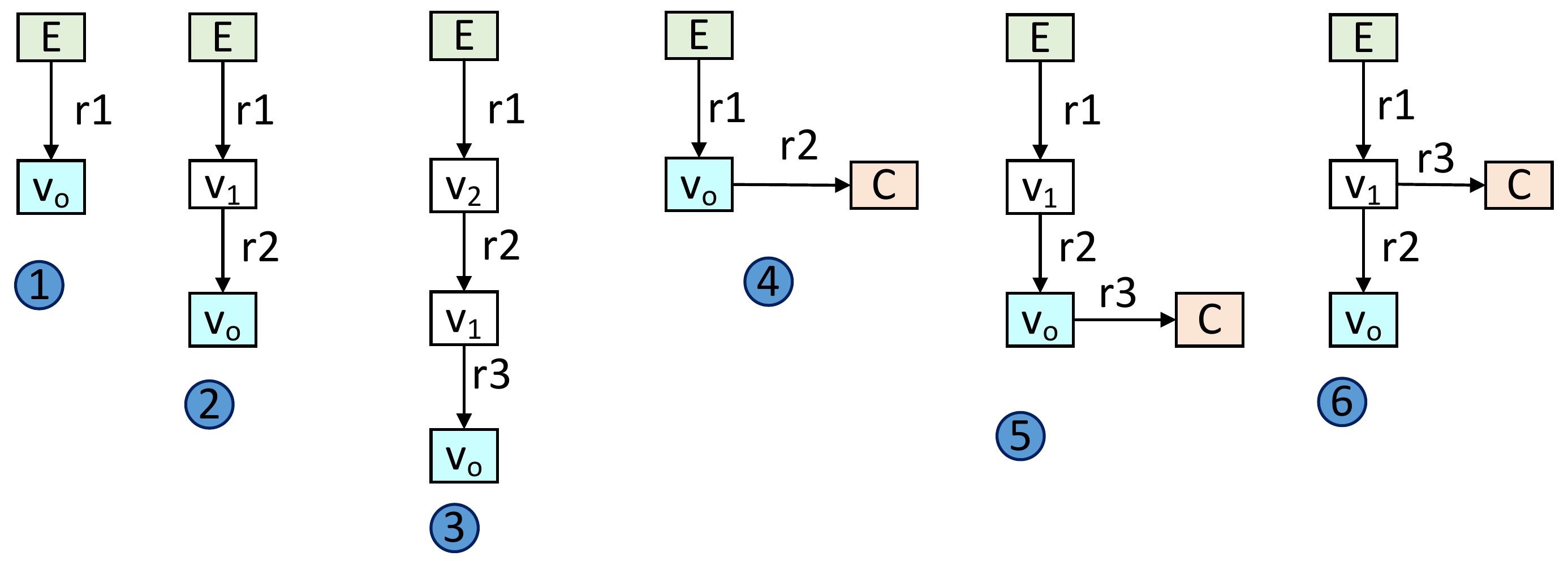}\vspace{-10pt}
	\caption{Six semantic structures defined in the paper. There are three semantic structures for questions in MetaQA: \textit{(SS1, SS2, SS3)}, and five semantic structures for questions in WSP: \textit{(SS1, SS2, SS4, SS5, SS6)}.}\label{fig:SS}
\end{figure}

\subsection{Semantic Structures}~\label{section:SS}
As observed by~\cite{chen2020formal}, if the structure of a question is known in advance, the noise in candidate query graphs can be reduced significantly by filtering. Thus, in this paper we define six semantic structures based on the question topology that is introduced by~\citet{srivastava-etal-2021-complex}. These six semantic structures are listed in Figure~\ref{fig:SS} and example questions for each semantic structure can be found in Figure~\ref{fig:excample SS} in Appendix. As we can see, a semantic structure is a graph that is an \emph{abstract} of the query graphs of the same pattern. Typically, a semantic structure consists of four components $\{E, r, v, C\}$, where $E$ denotes an entity, $r$ refers to all types of relations, $v$ is an existential variable, and $C$ denotes a constraint.

To identify the semantic structure of a question, we can train a classifier for prediction. But first we need to annotate each training question with its semantic structure. Fortunately, this annotation can be achieved readily for questions in MetaQA and WebQuestionsSP (WSP) since these questions are either partitioned by number of hops or accompanied by the SPARQL commands. Details on question annotation are provided in Sec.~\ref{Labeled SS} in Appendix. By annotating the questions in MetaQA and WSP, we found that these six semantic structures can cover 100\% of questions in MetaQA, and 77.02\% of questions in the test set of WSP as shown in Table~\ref{tab:coverage}. It is challenging to design additional semantic structures to cover 100\% of questions in WSP because there are some unusual operators in WSP, such as \texttt{Or} and \texttt{<=}, which are difficult to map to a common semantic structure. Even though there is only a 77.02\% coverage on the WSP test questions, our experiments show that SSKGQA already outperforms state-of-the-art methods on WSP. As a future work, we plan to explore new techniques to cover the remaining 22.98\% of questions in WSP. 

\begin{table}[t]
	\centering
	\renewcommand\arraystretch{1.3}
	\begin{tabular} {l|ccc|c}
		\hline 
		
		\hline
		\multirow{2}*{Dataset}& \multicolumn{3}{|c|}{MetaQA}  & \multirow{2}*{WSP}\\
		 & Hop-1 &Hop-2 & Hop-3 &  \\ 
		\hline		
		
		Train&100 & 100& 100 &91.37 \\
		Test&100 &100 & 100&77.02 \\ 
		Dev& 100& 100&100& n/a\\ 	
		\hline 
		
		\hline
	\end{tabular}
		\caption{Semantic structure coverage (\%) for questions in the training, test and development sets of MetaQA and WSP.}\label{tab:coverage}
\end{table}

\section{The Proposed Method}
We first provide an overview of SSKGQA, and then discuss its main components: (1) Structure-BERT and (2) query graph ranking in details.

\begin{figure}[t]  
	\centering
	\includegraphics[width=0.30\textheight]{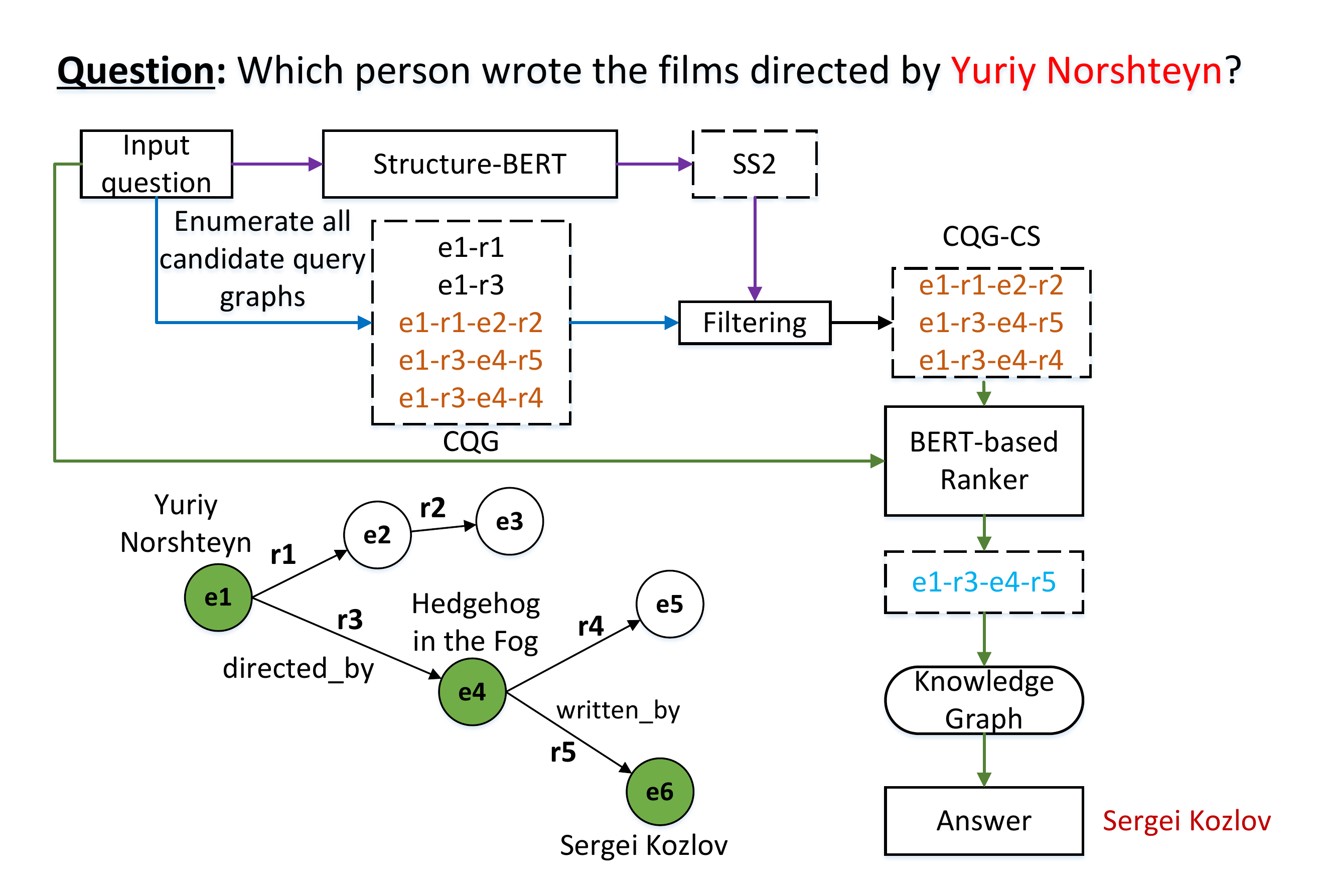}\vspace{-10pt}
	\caption{Overview of SSKGQA. A subgraph related to \texttt{Yuriy Norshteyn} from a KG is provided for illustration.}\label{fig:framework}  
\end{figure}

\subsection{Overview}
The overview of our proposed SSKGQA is depicted in Figure~\ref{fig:framework}. Given a question $q$, following previous works~\cite{Apoorv2020Improving,chen2020formal,cai2021deep} we assume the topic entity of $q$ has been obtained by preprocessing. Then the answer to $q$ is generated by the following steps. First, the semantic structure of $q$ is predicted by a novel Structure-BERT classifier. For the example in Figure~\ref{fig:framework}, $q$ is a 2-hop question and the classifier predicts its semantic structure as \textit{SS2}. Second, we retrieve all the candidate query graphs (CQG) of $q$ by enumeration\footnote{For clarify, only 1-hop and 2-hop candidate query graphs are considered in this example.}, and use the predicted semantic structure \textit{SS2} as the constraint to filter out noisy candidate query graphs and keep the candidate query graphs with correct structure (CQG-CS). Afterwards, a BERT-based ranking model is used to score each candidate query graph in CQG-CS, and the top-1 highest scored candidate is selected as the query graph $g$ for question $q$. Finally, the selected query graph is issued to KG to retrieve the answer \texttt{Sergei Kozlov}.

\subsection{Structure-BERT}
Given a question $q$, we first need to predict its semantic structure, which is a multi-class classification problem that classifies question $q$ to one of the six semantic structures defined in Figure~\ref{fig:SS}.

\begin{figure}[htbp]  
	\centering
	\includegraphics[width=0.30\textheight]{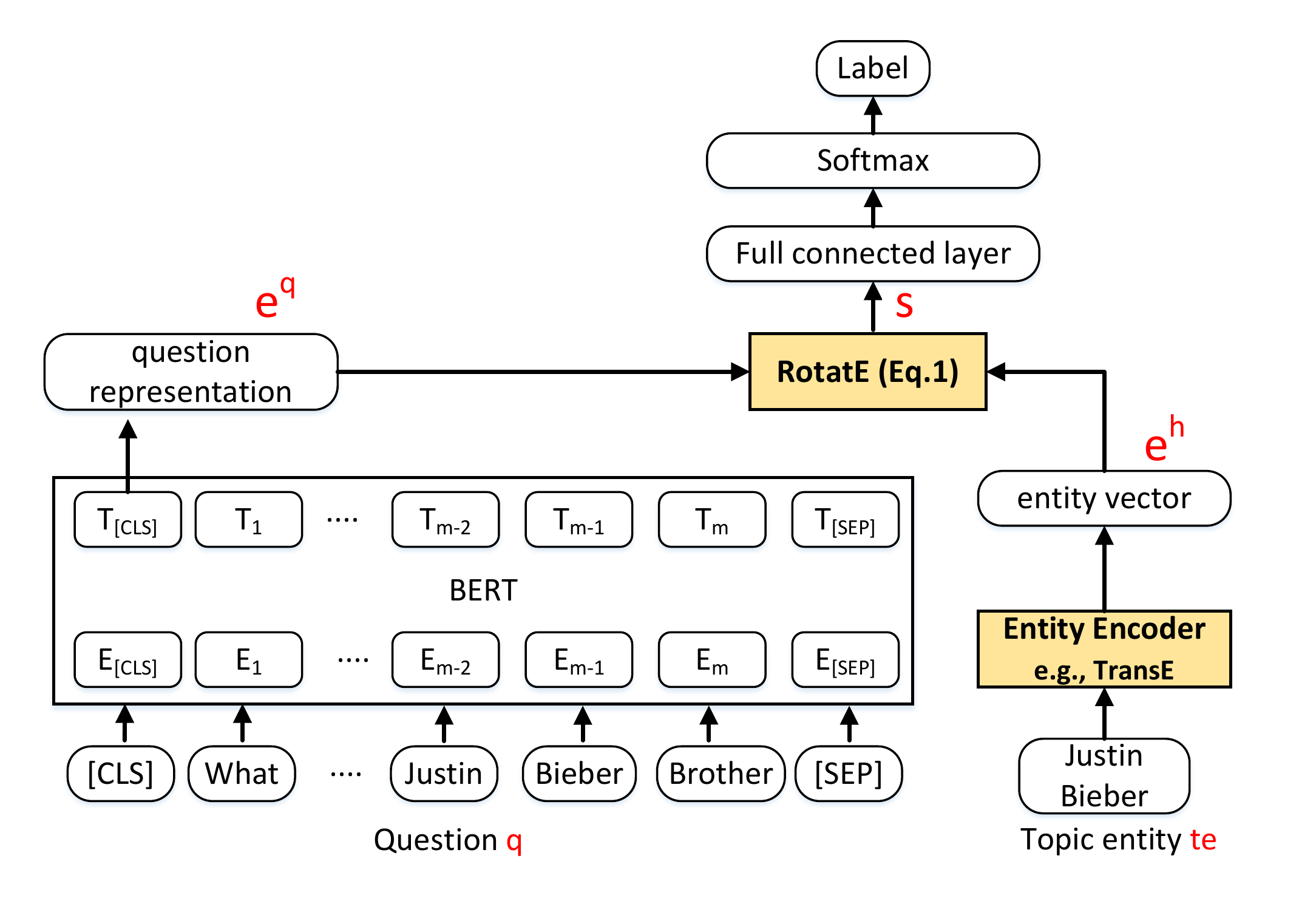}\vspace{-10pt}
	\caption{Structure-BERT: given a question and its topic entity, the model predicts the semantic structure of the question.}\label{fig:Structure-BERT}  
\end{figure}

Figure~\ref{fig:Structure-BERT} depicts the architecture of Structure-BERT. The input to Structure-BERT is question $q$ and its topic entity $te$. The output of Structure-BERT is a probability distribution over six semantic structures, i.e., $p(y|(q,te),\theta)$, where $\theta$ denotes the model parameters of Structure-BERT. Structure-BERT contains three sub-modules. \textbf{Question Encoder} encodes question $q$ by a BERT language model~\cite{devlin2018bert},
and the final hidden state corresponding to token \texttt{[CLS]} is used as the question embedding $e^q$. \textbf{Entity Encoder} leverages a pre-trained knowledge embedding model, such as
TransE~\cite{bordes2013translating} or ComplEx~\cite{trouillon2016complex}, to extract the entity embedding $e^h$. Next, the extracted question embedding $e^q$ and entity embedding $e^h$ are fed to a \textbf{RotatE} module for information fusion. First, we utilize a pre-trained RotatE~\cite{Sun2019RotatE} model to calculate a ``tail" embedding $e^t$ using $e^h$ and $e^q$, and then  fuse the topic entity, question and ``tail" embeddings by combining $e^h$, $e^q$ and $e^t$ to a latent representation $s$ by

\begin{align}
	&\bar{e}^h=e^h_{lh}+ie^h_{hh}\nonumber\\
	&\bar{e}^q=e^q_{lh}+ie^q_{hh}\nonumber\\
	&\bar{e}^t=\bar{e}^h\!\times\!\bar{e}^q\!=\!e^h_{lh}e^q_{lh}\!-\!e^h_{hh}e^q_{hh}\!+\!i(e^h_{hh}e^q_{lh}\!+\!e^h_{lh}e^q_{hh})\nonumber\\
	&s=e^h+e^q+e^t,
\end{align}
where $e_{lh}^*$ ($e_{hh}^*$) denotes the lower (higher) half of vector $e^*$. As such, $\bar{e}^*$ is a complex vector whose real (imaginary) part is from $e_{lh}^*$ ($e_{hh}^*$). Therefore, we can convert between $e^*$ and $\bar{e}^*$ readily. 

Finally, the latent representation $s$ is fed to a fully connected layer, followed by a softmax for classification. The whole network is fully differentiable and can be optimized by minimizing the traditional cross-entropy loss.

To train Structure-BERT, we need to annotate questions with their semantic structures to develop a training, validate and test set. As discussed in Sec.~\ref{section:SS}, this annotation can be conducted readily for questions in MetaQA and WSP. The details are provided in Sec.~\ref{Labeled SS} in Appendix.


\subsection{Query Graph Ranking}
Another important component of SSKGQA is a BERT-based ranking model for query graph ranking that can be trained by a triplet loss~\cite{facenet}. Specifically, the ranking model has three inputs: (1) question $q$=\{\texttt{[CLS]}, $w_1$, $w_2$, $\cdots$, $w_m$, \texttt{[SEP]}\}, where $w_i$ is the $i$-th word of $q$; (2) positive query graph\footnote{The positive query graph of a question can be found from the relation paths provided in MetaQA or the SPARQL command provided in WSP.} $g^p$=\{\texttt{[CLS]}, $u_1$, $u_2$, $\cdots$, $u_n$, \texttt{[SEP]}\}, where $u_i$ is the $i$-th unit of a query graph that is split by space or special symbol such as ``.". For example, given query graph (\texttt{Natalie Portman, film.actor.film, v}), $u_1$=\texttt{Natalie}, $u_2$=\texttt{Portman}, ... and $u_6$=\texttt{v}; and (3) negative query graph $g^n$ that is a candidate query graph in CQG-CS except the positive candidate of the question.

We utilize a BERT model $f(.)$ to extract the semantic representations $f(q)$, $f(g^p)$, $f(g^n)$ for $q, g^p, g^n$, respectively. This BERT model is built on a pre-trained BERT from Hugging Face\footnote{https://huggingface.co/bert-base-uncased}; we add one extra multi-head attention layer on top of the hidden state of the pre-trained BERT (See the ablation study in Sec.~\ref{sec:exp}). This BERT-based ranking model $f(.)$ is then optimized by minimizing the triplet loss~\cite{facenet}
\begin{align}
\max(\|f(q)\!\!-\!\!f(g^p)\|\!-\!\|f(q)\!\!-\!\!f(g^n)\|\!+\!\alpha,\!0),
\end{align}
where $\|.\|$ denotes the Euclidean distance and $\alpha$ is a margin parameter, which we set to 1 as default.

During training, the triplet loss reduces the distance between $f(q)$ and $f(g^p)$, while enlarging the distance between $f(q)$ and $f(g^n)$. At inference time, we calculate the similarity scores between question and its candidate query graphs from CQG-CS, and choose the top-1 highest scored candidate as query graph $g$ to retrieve final answer from KG.

\section{Experiments}\label{sec:exp}
We evaluate the performance of SSKGQA on two popular KGQA benchmarks: MetaQA and WebQuestionsSP (WSP), and compare it with seven state-of-the-arts methods. Ablation study is also conducted to understand the effectiveness of different components of SSKGQA. 

Our PyTorch source code is provided at \url{https://github.com/ToneLi/SSKGQA}. All our experiments are performed on Nvidia RTX GPUs. 

\subsection{Datasets}
\begin{itemize}
	\item \textbf{MetaQA}~\cite{Zhang2017Variational} is a large scale KGQA dataset with more than 400k questions. It contains questions with 1, 2 or 3 hops. In our experiments, we use the vanilla version of the QA dataset. MetaQA also provides a KG from the movie domain with 43,233 entities, 9 relations and 134,741 triples.

	\item \textbf{WebQuestionsSP (WSP)}~\cite{Yih2016The} is a small scale KGQA dataset with 5,119 questions which are answerable through Freebase KG. Since Freebase has more than 338,580,000 triples, for ease of experimentation we use a light version provided by ~\citet{Apoorv2020Improving}. This smaller KG has 1.8 million entities and 5.7 million triples.
\end{itemize}
The statistics of training, development and test datasets of MetaQA and WSP is provided in Table~\ref{tab:statistics}. Compared to MetaQA, WSP is relatively small QA dataset even though its KG is much larger than that of MetaQA.

\begin{table}[h]
	\centering
	\renewcommand\arraystretch{1.3}
	\begin{tabular} {lrrr}
		\hline 
		
		\hline
		Dataset& Train & Dev  & Test   \\ 
		\hline		
		MetaQA- hop1&96,106 & 9,992& 9,947  \\
		MetaQA- hop2& 118,980 & 14,872 & 14,872 \\ 
		MetaQA- hop3& 114,196& 14,274&14,274\\ 	
		\hline 
		WSP& 3,304 & -- & 1,815\\ 
		\hline 
		
		\hline
	\end{tabular}
		\caption{Statistics of the MetaQA and WSP datasets}\label{tab:statistics}
\end{table}

\subsection{Hyperparameter Settings}
\paragraph{Structure-BERT} 

We set the dropout rate to 0.1, batch size to 32, and use AdamW optimizer~\cite{loshchilov2017decoupled} with the learning rate of 5e-8. We also apply gradient clipping to constrain the maximum value of $L_2$-norm of the gradients to be 1. To extract the latent representations of topic entities, pre-trained ComplEx~\cite{trouillon2016complex} and TransE~\cite{bordes2013translating} are adopted for MetaQA and WSP, respectively. 

\paragraph{BERT-based Ranking Model} We add one extra multi-head attention layer on top of the hidden state of the pre-trained BERT. This extra multi-head attention layer contains three attention heads and a 3072-dim fully connected layer. The dropout rate is set to 0.5. We use AdamW Optimizer~\cite{loshchilov2017decoupled} with the learning rate of 2e-5. We also use gradient clipping to constrain the max $L_2$-norm of the gradients to be 1.  

\subsection{Baselines} 
We compare our SSKGQA against seven state-of-the-art complex KGQA models: 1) GraftNet~\cite{Sun2018Open}, which answers the questions based on the subgraphs it creates. 2) PullNet~\cite{Sun2019PullNet}, which proposes a ``pull" operation to retrieve the relevant information from KG and external corpus. 3) Key-Value Memory Network (KV-MemNN)~\cite{Miller2016Key}, which uses key-value pairs as the memory units to answer questions. 4) EmbedKGQA~\cite{Apoorv2020Improving}, which proposes a knowledge embedding method for Complex KGQA. 5) TransferNet~\cite{shi2021transfernet}, which utilizes an interpretable model for complex KGQA. 6) DCRN~\cite{cai2021deep}, which proposes a Bayesian network to retrieve the final answers. For MetaQA, we also include 7) VRN~\cite{Zhang2017Variational} as the baseline, which proposes an embedding reasoning graph and utilizes variational inference to improve the performance of Complex KGQA. 

\begin{table}
	\centering
	\renewcommand\arraystretch{1.1}		\begin{tabular} {lcccc}
        \hline

		\hline
		Model& Hop-1 & Hop-2  & Hop-3 &  WSP  \\ 
		\hline		
		KV-MemNN& 96.2 & 82.7& 48.9&46.7\\ 
		VRN&97.5& 89.2& 62.5&- \\ 
		GraftNet& 97.0&94.8& 77.7&66.4 \\ 
		PullNet& 97.0&99.9&91.4&68.1\\ 		
		EmbedKGQA& 97.5 & 98.8 & 94.8&66.1\\ 
		DCRN& 97.5 & 99.9 & 99.3&67.8\\
    		TransferNet& 97.5 &\textbf{100} & \textbf{100}&\textbf{71.4}\\ 
		\hline 
		SSKGQA& \textbf{99.1} & 99.7 &99.6&\textbf{71.4}\\ 
		\hline 
		
		\hline
	\end{tabular}
		\caption{Hits@1 values of different KGQA methods on MetaQA and WSP. Hop-$n$ denotes the hop-$n$ questions of MetaQA.}
	\label{Ref:hit1 inMetaQA and WSP}
\end{table}	

\subsection{Comparison with State-of-the-Arts} Table~\ref{Ref:hit1 inMetaQA and WSP} reports the performances of SSKGQA and seven state-of-the-art methods on MetaQA and WSP. As can be seen, the performances of KV-MemNN are limited by the error propagation over multi-hop reasoning, i.e., as the number of hops increases, its performance is degraded significantly. GraftNet and PullNet perform similarly well on all datasets (expect MetaQA-hop3) as both of them rely on subgraphs to retrieve the answers. Compared to GraftNet, PullNet has much improved results on MetaQA-hop3, indicating that the proposed pull operation is more suitable to complex questions. EmbedKGQA achieves a good performance on MetaQA, but a relatively lower performance on WSP. This is because treating question as a relation path in a triple may introduce more noise especially when the question is more complex. Even though DCRN achieves the best performance on MetaQA-hop2, it still suffers from error propagation when inferring the reasoning paths. For more complex WSP questions, DCRN has a 3.6-point lower accuracy than that of our method. In general, TransferNet is the most competitive one to our SSKGQA. While both methods have the best results on WSP, SKGQA has an improved performance on MetaQA-hop1 over TransferNet,  and is almost neck to neck on hop-2 and hop-3. Overall, SSKGQA outperforms or achieves comparable exact-match hits@1 performances to the other methods, demonstrating the effectiveness of our proposed method. 

\subsection{Ablation Study}
We further investigate the effectiveness of different components of SSKGQA, including semantic structure based filtering, Structure-BERT, the BERT-based ranking model, etc. 

\subsubsection{Impact of Semantic Structure based Filtering}

One of the core ideas of SSKGQA is the semantic structure based filtering. In this section, we evaluate the effectiveness of this operator by enabling / disabling it and report the final performances of SSKGQA, which correspond to the w/ SS and w/o SS results in Table~\ref{ref:with ss}. For the purpose of illustration, when we enable the filtering (w/ SS), we assume that our Structure-BERT classifier can correctly predict the semantic structures of all the questions with a 100\% accuracy, and therefore the impact of the filtering isn't affected by the accuracy of the classifier. For ease of experimentation, we use a BiGRU as the ranking model in this experiment.

Table~\ref{ref:with ss} reports the impacts of the semantic structure based filtering. It can be observed that for simple questions, e.g., MetaQA-hop1, SSKGQA w/ SS and w/o SS have very similar performances. However, when the questions are more complex, SSKGQA w/ SS achieves significantly higher accuracies (sometimes over 10\%) than SSKGQA w/o SS, demonstrating the effectiveness of the semantic structure based filtering for complex questions.

\begin{table}[t]
	\centering
	\renewcommand\arraystretch{1.1}
	\begin{tabular} {lllll }
		\hline
		
		\hline 
		&Hop-1& Hop-2 & Hop-3&  WSP\\ 
		\hline		
		w/o SS&99.11& 93.71 & 62.10&  45.89\\ 
		\hline 
		w/ SS&\textbf{ 99.26}&\textbf{ 99.03} & \textbf{95.69}& \textbf{ 58.51}\\ 
		\hline
		
		\hline
	\end{tabular}
		\caption{Hits@1 values of SSKGQA w/ SS and w/o SS on MetaQA and WSP.}
	\label{ref:with ss}
\end{table}

\subsubsection{Accuracy of Structure-BERT}
Structure-BERT plays a critical role in SSKGQA as it predicts the semantic structure of a question, which is then used to filter out noisy candidate query graphs. In this section, we evaluate the accuracy of Structure-BERT and compare it with other design choices.

Specifically, we compare the performance of Structure-BERT with four other classifiers, including BiGRU and three pre-trained language models: BERT~\cite{devlin2018bert}, DistilBERT~\cite{budzianowski2019hello} and CamemBERT~\cite{martin2020camembert}). For these four classifiers, they classify a question directly to one of the six semantic structures without considering topic entity and information fusion as in Structure-BERT. 

\begin{table}[h]
	\centering
	\renewcommand\arraystretch{1.1}
	\begin{tabular} {lllll}
		\hline
		
		\hline 
		Model&   Hop1&Hop2&Hop3&WSP\\ 
		\hline		
		BiGRU&  96.44&94.49&98.83&80.95\\
		BERT&  94.52&98.70&96.22&82.62\\
		DistilBERT&  95.66&98.30&97.02& 83.37\\
		CamemBERT&  96.66&97.30&98.26& 81.90\\
		Structure-BERT&\textbf{99.24}  &\textbf{99.87}&\textbf{99.73}&\textbf{86.97}\\ 
		\hline 
	
		\hline
	\end{tabular}
	\caption{Classification accuracies of different classifiers on predicting semantic structures of questions from MetaQA and WSP.}		\label{ref:structure:bert}
\end{table}

Table~\ref{ref:structure:bert} reports the classification accuracies of different classifiers on the questions from MetaQA and WSP. As can be seen, our Structure-BERT achieves nearly 100\% accuracies on MetaQA and 86.97\% accuracy on WSP, demonstrating the effectiveness of Structure-BERT on semantic structure prediction. Further, Structure-BERT consistently outperforms all the other classifiers by notable margins, indicating the importance of leveraging both question and topic entity for information fusion for semantic structure prediction. We also notice that the classification accuracy on WSP is much lower than that of MetaQA. This is likely due to: (1) the class imbalance issue of the WSP questions, and (2) much smaller number of training questions in WSP (3,304) than that of MetaQA (329,282). We will leave the further improvements of Structure-BERT on WSP to future works.

\subsubsection{Performance of BERT-based Ranking Model}

The BERT-based ranking model decides which candidate query graph is to be used to retrieve the final answer. Therefore, its performance is of the paramount importance to SSKGQA. In this section, we evaluate the effectiveness of our proposed BERT-based ranking model and compare it with other three ranking methods, including 1) CNN~\cite{yih2015semantic}, which uses a CNN to learn the representation of question and candidate query graph for ranking. 2) BiGRU, which uses a BiGRU to learn the representation of question and candidate query graph for ranking.
3) BERT~\cite{devlin2018bert}, which uses a pre-trained BERT\footnote{https://huggingface.co/bert-base-uncased} to extract the representation of question and candidate query graph for ranking. 


\begin{table}[h]
	\centering
	\renewcommand\arraystretch{1.1}
	\begin{tabular} {lllll}
		\hline 
		
		\hline
		Model &Hop-1&Hop-2&Hop-3&WSP\\ 
		\hline		
		CNN   &97.70 &99.21 &92.91&50.24\\ 
		BiGRU &98.87 &98.95 &95.43&56.51\\  
		BERT &\textbf{99.49} &99.26 &99.54& 71.02\\ 
	    BERT$^*$ (ours) &99.10 &\textbf{99.69} &\textbf{99.64}&\textbf{71.40}\\  
		\hline
		
		\hline
	\end{tabular}
		\caption{Hits@1 values of different ranking models on MetaQA and WSP. BERT$^*$ denotes our BERT-based ranker.}
	\label{tab:ranking model}
\end{table}

Table~\ref{tab:ranking model} reports the performances of different ranking models on MetaQA and WSP, where BERT$^*$ denotes our proposed BERT-based ranking model that has one extra multi-head attention layer on top of the hidden state of the pre-trained BERT. As we can see, the BERT-based ranking models (BERT and BERT$^*$) outperform the traditional CNN or BiGRU based ranking models since the former can leverage large scale pre-trained BERT for transfer learning. Our BERT$^*$ further improves the performance of the pre-trained BERT due to the additional attention layer which enables model to reweight the attention values to different semantic units in the input and enhance the semantic representation of question and candidate query graphs for ranking.

To validate the design choices of our BERT-based model, we run additional ablation studies on different factors of our ranking model, such as number of negative query graphs for the triplet loss based training and number of heads in the added multi-head attention layer. The details are relegated to Sec.~\ref{ref:more_ablation} in Appendix.

\section{Conclusions}

This paper introduces SSKGQA, a semantic structure based method to predict query graphs from natural language questions. Compared to prior query graph prediction based methods, SSKGQA filters out noisy candidate query graphs based on the semantic structures of input questions. To this end, we define six semantic structures from common questions of the KGQA benchmarks. A novel Structure-BERT classifier is then introduced to predict the semantic structure of each question, and a BERT-based ranking model with a triplet loss is proposed for query graph ranking. Extensive experiments on MetaQA and WSP demonstrate the superior performance of SSKGQA over seven state-of-the-art methods.

As for future work, we plan to investigate techniques to design additional semantic structures to cover the remaining 22.98\% of questions in WSP. We would like also to improve Structure-BERT's accuracy on WSP by addressing the class imbalance and data scarcity issues of WSP.

\section*{Acknowledgements}
We would like to thank the anonymous reviewers for their comments and suggestions,
which helped improve the quality of this paper. We would also gratefully acknowledge
the support of VMware Inc. for its university research fund to this research.

\bibliographystyle{acl_natbib}
\bibliography{custom}


\section{Appendix}~\label{Appendices}
\subsection{Example Questions and their Semantic Structures}
\label{SS excample}
We provide six example questions and their corresponding semantic structures. These examples are selected from the MetaQA and WSP benchmarks. $E$ denotes a topic entity, $r$ refers to all types of relations, $v$ is an existential variable, and $C$ denotes a constraint.

\begin{figure}[htbp]  
	\centering
	\includegraphics[width=0.3\textheight]{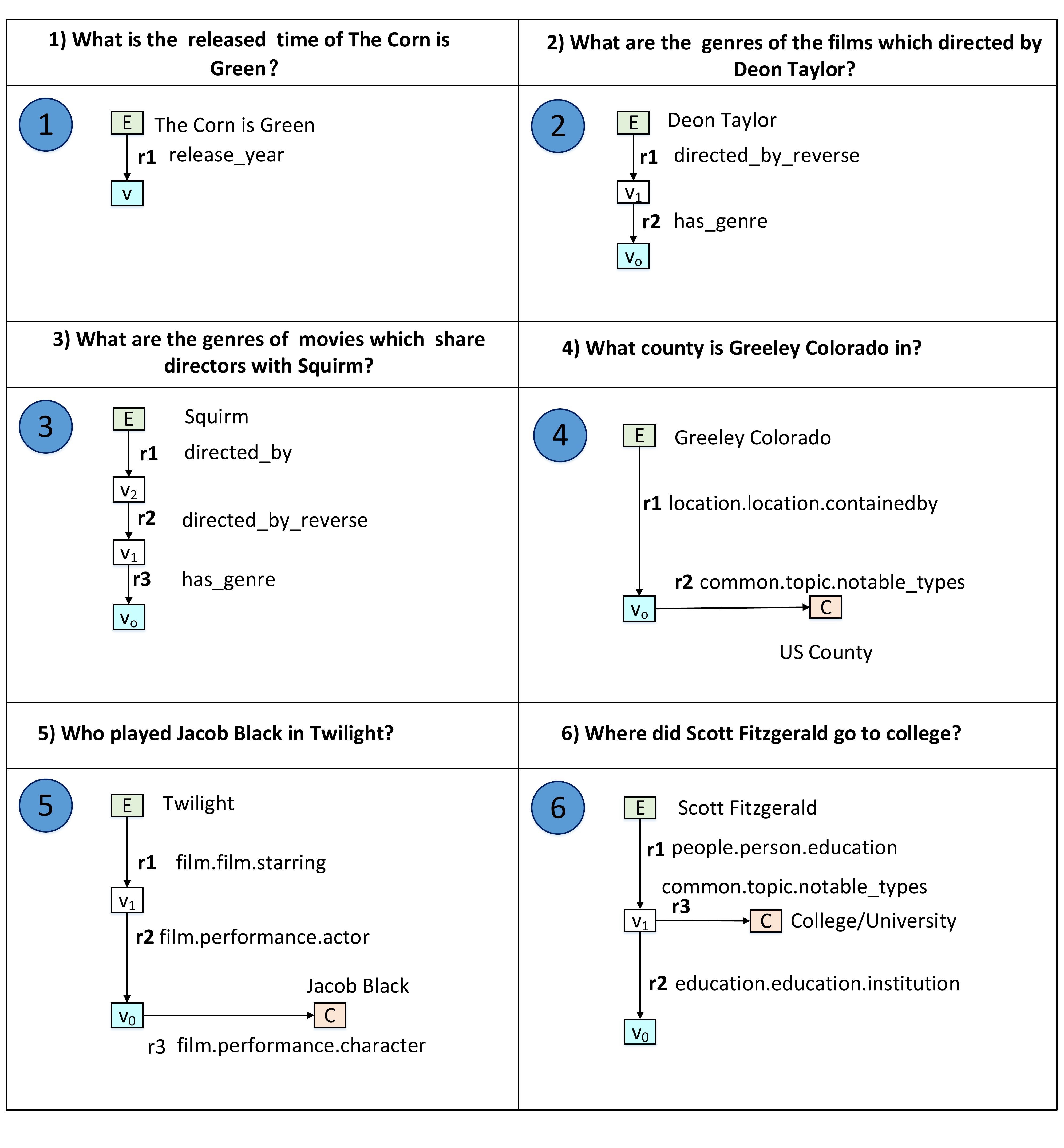}\vspace{-10pt}
	\caption{Examples questions and their semantic structures}\label{fig:excample SS}  
\end{figure}

\subsection{Semantic Structure Annotation for Training Questions}
\label{Labeled SS}
We need to annotate each training question to a semantic structure in order to train Structure-BERT for semantic structure prediction. Here we describe how we can automatically annotate each training question in MetaQA and WSP.

\subsubsection{MetaQA} 
Annotating semantic structure for each training question in MetaQA is straightforward. Since training questions are organized by number of hops, all training questions with 1-hop are labeled to SS1, the 2-hop ones to SS2, and the 3-hop ones to SS3, respectively. 

\subsubsection{WSP} 
\begin{figure}[htbp]  
	\centering
	\includegraphics[width=0.3\textheight]{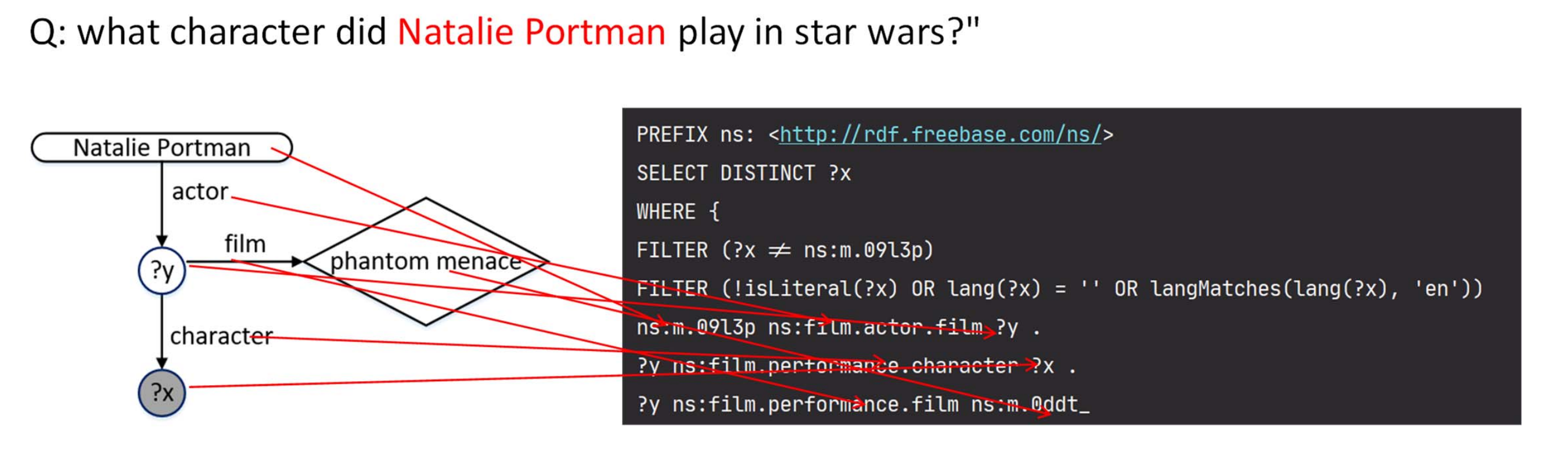}\vspace{-10pt}
	\caption{Mapping from a SPARQL command to its query graph}\label{fig:SS genation}  
\end{figure}

The SPARQL command for each question is provided in WSP. Thus, we can readily extract the query graph of a question from its SPARQL command. See an example in Figure~\ref{fig:SS genation}, where a query graph is shown on the left and the corresponding SPARQL command is shown on the right; the correspondences between two parts are marked by red lines. Once the query graph is extracted, each training question can be readily annotated to a semantic structure based on its query graph.

\subsection{Additional Ablation Study}\label{ref:more_ablation}
\begin{table}[h]
	\centering
	\renewcommand\arraystretch{1.1}
	\begin{tabular} {lllll}
		\hline
		
		\hline
		Number of Neg. & 1&5&10&50\\ 	
		hits@1 & 64.57&68.87&70.08&70.79\\ 
		\hline	
		Number of Neg. &100&200&300&500\\
		hits@1& \textbf{71.29}&71.23 &\textbf{71.29}&70.58\\
		\hline
		
		\hline
	\end{tabular}

	\caption{Hits@1 values on WSP of our BERT-based ranking model when trained with different number of negative query graphs.}
		\label{ref:query graph analysis}
\end{table}
\paragraph{Number of negative query graphs} Given a question, a plenty of negative query graphs can be generated by enumeration from a KG. By analyzing the questions in WSP, we found that the maximal number of negative query graphs that can be extracted for WSP is around 500. We need to determine a proper number of negative samples to train the ranking model with the triplet loss. To this end, we evaluate the performances of our BERT-based ranker when trained with different numbers of negative samples. The results are reported in Table~\ref{ref:query graph analysis}, where 8 different number of negative samples are considered. As we can see, when number of negative samples is over 100, the ranking model achieves improved hits@1 performances; when $n=100$ our BERT-based ranking model yields the best hits@1 with a good run-time performance.

\paragraph{Number of heads} 
\begin{table}[h]
	\centering
	\renewcommand\arraystretch{1.3}
	\begin{tabular} {lccc}
		\hline
		
		\hline
		Number of Heads& 1&3&6\\ 
		\hline	
		hits@1 &71.29 &\textbf{71.40} &70.63\\ 
		\hline
		
		\hline
	\end{tabular}
	
\label{ref:head number}
	\caption{Hits@1 values on WSP of our BERT-based ranking model with different number of attention heads in the added attention layer.}\label{tab:number of heads}
\end{table}

Another design choice for the extra multi-head attention layer in our BERT-based ranker is the number of attention heads. We therefore evaluate the performance of our BERT-based ranker with 3 different number of attention heads. The results in Table~\ref{tab:number of heads} shows that when the number of attention heads is 3, our BERT-based ranker achieves the best hits@1.



\end{document}